\pdfoutput=1

\documentclass[11pt]{article}

\usepackage[preprint]{acl}

\usepackage{times}
\usepackage{latexsym}

\usepackage[T1]{fontenc}

\usepackage[utf8]{inputenc}

\usepackage{microtype}

\usepackage{inconsolata}

\usepackage{graphicx}
\usepackage{array,multirow}
\usepackage{dirtytalk}
\usepackage{stfloats}

\patchcmd{\quote}{\rightmargin}{\leftmargin 1.25em \rightmargin}{}{}

\title{Thinking Outside of the Differential Privacy Box: \\ A Case Study in Text Privatization with Language Model Prompting}

\author{Stephen Meisenbacher \and Florian Matthes \\
Technical University of Munich\\
School of Computation, Information and Technology \\
Department of Computer Science\\
Garching, Germany\\
\texttt{\{stephen.meisenbacher,matthes\}@tum.de} \\
}

\begin{document}
\maketitle
\begin{abstract}
The field of privacy-preserving Natural Language Processing has risen in popularity, particularly at a time when concerns about privacy grow with the proliferation of Large Language Models. One solution consistently appearing in recent literature has been the integration of Differential Privacy (DP) into NLP techniques. In this paper, we take these approaches into critical view, discussing the restrictions that DP integration imposes, as well as bring to light the challenges that such restrictions entail. To accomplish this, we focus on \textsc{DP-Prompt}, a recent method for text privatization leveraging language models to rewrite texts. In particular, we explore this rewriting task in multiple scenarios, both with DP and without DP. To drive the discussion on the merits of DP in NLP, we conduct empirical utility and privacy experiments. Our results demonstrate the need for more discussion on the usability of DP in NLP and its benefits over non-DP approaches. 
\end{abstract}

\section{Introduction}
The topic of privacy in Natural Language Processing has recently gained traction, which has only been fueled by the prominent rise of Large Language Models. In an effort to address concerns revolving around the protection of user data, the study of privacy-preserving NLP has presented a plethora of innovative solutions, all investigating in some form the optimization of the privacy-utility trade-off for the safe processing of textual data.

A well-studied solution comes with the integration of Differential Privacy (DP) \cite{dwork2006differential} into NLP techniques. Essentially, the use of DP entails the addition of calibrated noise to some stage in a pipeline, e.g., directly to the data or to model weights. This is performed with the ultimate goal of protecting the individual whose data is being used, aligned with the objective of Differential Privacy set out in its inception nearly 20 years ago.

The incentive of proving Differential Privacy is the mathematical guarantee of privacy protection that it offers, so long as its basic principles are adhered to. Particularly, important DP notions must be strictly defined, such as who the individual is, how data points are adjacent, and how data can be bounded. As such, the fusion of Differential Privacy and NLP introduces several challenges \cite{feyisetan2021research, habernal-2021-differential, klymenko-etal-2022-differential, mattern-etal-2022-limits}. 
When generalized forms of DP are used or well-defined notions of DP concepts are lacking, the promise of DP becomes more of a shallow guarantee.

In this work, we critically view the pursuit of DP in NLP, focusing on the particular method of \textsc{DP-Prompt} \cite{Utpala2023LocallyDP}. This method leverages generative Language Models to rewrite (paraphrase) texts with the help of a DP token selection method based on the Exponential Mechanism \cite{mattern-etal-2022-limits}. We run experiments on three rewriting settings: (1) \textit{DP}, (2) \textit{Quasi-DP}, and (3) \textit{Non-DP}; the purpose of this trichotomy is to explore the benefits and shortcomings of DP in text rewriting. We define our research question as:

\begin{quote}
    \vspace{-5pt}
    \textit{What is the benefit of integrating Differential Privacy into private text rewriting methods leveraging LMs, and what effect can be observed by relaxing this guarantee?}
    \vspace{-5pt}
\end{quote}

Our empirical findings show the advantages that incorporating DP into text rewriting mechanisms brings, notably higher semantic similarity and resemblance to the original texts, along with strong empirical privacy results. This, however, comes with the downside of generally lower quality text in terms of readability, particularly at stricter privacy budgets. These findings open the door to discussions regarding the practical distinction between DP and non-DP text privatization, where we present open questions and paths for future work.

The contributions of our work are as follows:
\begin{enumerate}
    \itemsep -0.1em
    \item We explore the merits of DP in LM text rewriting through comparative experiments.
    \item We evaluate \textsc{DP-Prompt} in a series of utility and privacy tests, and analyze the difference in DP vs. non-DP privatization.
    \item We call into question the merits of DP in NLP, presenting the benefits and limitations of doing so as opposed to non-DP privatization.
\end{enumerate}


\section{Related Work}
\label{sec:related}
Natural language can leak personal information \cite{Brown2022WhatDI} and it is possible to extract training data from Machine Learning models \cite{Pan2020, carlini2021extracting, mattern-etal-2023-membership}. In the \textit{global} DP setting, user texts are collected at a central location and a model is trained using privacy-preserving optimization techniques \cite{ponomareva-etal-2022-training, Kerrigan2020DifferentiallyPL} such as DP-SGD \cite{Abadi2016DeepLW}. 
The primary drawback of this model is that user data must be collected at a central location, giving a data curator access to the entire data \cite{klymenko-etal-2022-differential}. To mitigate this, text can be obfuscated or rewritten locally in a DP manner before collecting it at a central location \cite{feyisetan_balle_2020, Igamberdiev.2023.ACL, hu-etal-2024-differentially}.

The earliest set of approaches of DP in NLP began at the word level \cite{weggenmann2018syntf, fernandes2019generalised, yue, Chen2022ACT, carvalho2023tem, meisenbacher20241}, yet these methods do not consider contextual and grammatical information during privatization \cite{mattern-etal-2022-limits, meisenbacher2024comparative}. Other works operate directly at the sentence level by either applying DP to embeddings \cite{Meehan2022SentencelevelPF} or latent representations \cite{Bo2019ERAEDP, 10.1145/3485447.3512232, Igamberdiev.2023.ACL}. DP text rewriting methods using generative LMs \cite{mattern-etal-2022-limits, Utpala2023LocallyDP, flemings-annavaram-2024-differentially} or encoder-only models \cite{meisenbacher-etal-2024-dp} have also been proposed.

\section{Method}
\label{sec:method}
Here, we describe the base text privatization method that we utilize, as well as the variations which form the basis of our experiments.

\subsection{\textsc{DP-Prompt}}
\textsc{DP-Prompt} \cite{Utpala2023LocallyDP} is a differentially private text rewriting method in which users generate privatized documents at the local level by prompting Language Models to rewrite input texts. In particular, the LMs are prompted to \textit{paraphrase} a given text. The immediate advantage of this method comes with the flexibility in model choice as well as the generalizability to all general-purpose pre-trained (instruction-finetuned) LMs.

The integration of DP into this rewriting process comes at the generation step, where for each output token, a DP token selection mechanism is implemented in the form of \textit{temperature sampling}. In \citet{mattern-etal-2022-limits}, it is shown that the use of temperature can be equated to the Exponential Mechanism \cite{exponential-mechanism}. Relating this mechanism to the privacy budget $\varepsilon$ of DP, the authors show that $\varepsilon = \frac{2\Delta}{T}$, where $T$ is the temperature and $\Delta$ is the \textit{sensitivity}, or range, of the token logits. A fixed sensitivity can be ensured by \textit{clipping} the logits to certain bounds.


For the purposes of this work, we perform all experiments using \textsc{DP-Prompt} with the \textsc{flan-t5-base} model from Google \cite{chung2022scaling}.

\subsection{Rewriting Approaches}
Motivated by the \textsc{DP-Prompt} rewriting mechanism, we introduce three privatization strategies based on its DP token selection mechanism:
\begin{enumerate}
    \itemsep -0.1em
    \item \textbf{DP}: we use \textsc{DP-Prompt} as originally introduced, namely by clipping logit values and scaling logits by temperatures calculated based on $\varepsilon$ values. We test on the values $\varepsilon \in \{25, 50, 100, 150, 250\}$. Logits are clipped based on an empirical measurement of logits in the \textsc{flan-t5-base} model\footnote{Specifically, to the range ($logit\_mean$, $logit\_mean + 4\cdot logit\_std$) = (-19.23, 7.48), thus $\Delta = 26.71$.}.
    \item \textbf{Quasi-DP}: we replicate the \textbf{DP} strategy \textit{without} clipping, i.e., only using temperature sampling based on the abovementioned $\varepsilon$ values. We call this \textit{quasi-DP} since the temperature values $T$ are calculated \textit{as if} clipping was performed (i.e., sensitivity is bounded), but the \textit{unbounded logit range} is actually used.
    \item \textbf{Non-DP}: here, we do not use any clipping or temperature, but rather only vary the \textit{top-k} parameter, or the number $k$ of candidate tokens considered when sampling the next token. We choose $k \in \{50, 25, 10, 5, 3\}$.
\end{enumerate}

With these three privatization strategies, we aim to measure empirically the effect on utility and privacy by strictly enforcing DP, relaxing DP, and by performing privatization devoid of DP. In this way, one may be able to analyze the merits of DP-based text privatization methods, and furthermore, observe the theoretical guarantees of DP in action.

\section{Experimental Setup and Results}
\label{sec:setup}
As stated by \citet{mattern-etal-2022-limits}, a practical text privatization mechanism should: (1) protect against deanonymization attacks, (2) preserve utility, and (3) keep the original semantics intact. As such, we design our experiments by leveraging multiple dimensions of a single dataset. The results of all described experiments can be found in Table \ref{tab:results}.

\subsection{Dataset}
For all of our experiments, we utilize the Blog Authorship Corpus \cite{schler2006effects}. This corpus contains nearly 700k blog post texts from roughly 19k unique authors. 
The corpus also lists the ID, gender, and age of author for each blog post. Full details on the preparation of the corpus are found in Appendix \ref{sec:dataset}; pertinent details are outlined below.

We prepare two subsets of the corpus. The first, which we call \textbf{author10}, only considers blog posts from the top-10 most frequently occurring blog authors in the corpus. This subset results in a dataset of 15,070 blog posts spanning five categories.

The second subset, called \textbf{topic10}, is necessary as the classification of the gender and age attributes for the \textbf{author10} dataset would be a less diverse and challenging task. We first take a random 10\% sample of the top-10 topics from the filtered corpus, resulting in a sample of 14,259 blogs. Here, the \textit{age} value is binned into one of five bins to ensure an equal number of instances in each bin.

\subsection{Utility Experiments}
We perform utility experiments for both the \textbf{author10} and \textbf{topic10} datasets. To measure utility across all privatization strategies, we first privatize each dataset on all selected privatization parameters. As we choose 5 parameters ($\varepsilon$/$T$ or $k$) for each of our three strategies, this results in 15 dataset variants, i.e., 15 results per metric, each of which represents the average between the two datasets.

\paragraph{Semantic Similarity.}
To measure the ability of each privatization strategy to preserve the semantic meaning of the original sentence, we employ two similarity metrics: BLEU \cite{10.3115/1073083.1073135} and cosine similarity. Both metrics strive to capture the similarity between output (in this case privatized) text and a reference (original) text; BLEU relies on token overlap while cosine similarity between embeddings is more contextual.

We use SBERT \cite{DBLP:journals/corr/abs-1908-10084} to calculate the average cosine similarity (CS) between the original blog posts and their privatized counterparts. For this, we use utilize three embeddings models to account for model-specific differences: \textsc{all-MiniLM-L6-v2}, \textsc{all-mpnet-base-v2}, and \textsc{gte-small} \cite{li2023general}. For each dataset, we report the mean of the average cosine similarity calculated for each model.

We also report the BLEU score between privatized texts and their original counterparts. This is done using the BLEU implementation made available by Hugging Face. As before, reported BLEU scores are the average across an entire dataset.

\paragraph{Readability.}
In addition, we also measure the quality and readability of the privatized outputs by using perplexity (PPL) \cite{10.1145/3485447.3512232}, specifically with GPT-2 \cite{radford2019language}.

\begin{table*}[htbp]
\resizebox{\textwidth}{!}{
    \begin{tabular}{l|c|ccccc|ccccc|ccccc}
 & Baseline & \multicolumn{5}{c|}{DP} & \multicolumn{5}{c|}{Quasi-DP} & \multicolumn{5}{c}{Non-DP} \\ \hline
$\varepsilon$/$k$ value & $\infty$ & 25 & 50 & 100 & 150 & 250 & 25 & 50 & 100 & 150  & 250 & 50 & 25 & 10 & 5 & 3\\ \hline
CS $\uparrow$ & 1.00 & 0.589 & 0.597 & 0.812 & 0.827 & 0.832 & 0.347 & 0.598 & 0.810 & 0.826 & 0.833 & 0.710 & 0.726 & 0.750 & 0.741 & 0.787 \\
 BLEU $\uparrow$ & 1.00 & 0.077 & 0.029 & 0.123 & 0.142 & 0.153 & 0.001 & 0.029 & 0.121 & 0.141 & 0.153 & 0.049 & 0.054 & 0.063 & 0.063 & 0.088 \\
PPL $\downarrow$ & 41 & 8770 & 1234 & 928 & 919 & 905 & 16926 & 1380 & 982 & 932 & 925 & 816 & 972 & 1080 & 827 & 837 \\ \hline \hline
Author F1 (s) $\downarrow$ & 66.45 & 7.13 & 37.05 & 58.10 & 61.12 & 60.60 & 6.59 & 36.91 & 57.84 & 60.37 & 61.13 & 46.83 & 47.07 & 49.88 & 51.69 & 53.10 \\
Author F1 (a) $\downarrow$ & 66.45 & 2.68 & 33.52 & 52.82 & 55.46 & 57.35 & 2.74 & 33.29 & 54.81 & 55.64 & 57.34 & 42.56 & 44.81 & 45.20 & 48.22 & 49.91 \\
Gender F1 (s) $\downarrow$ & 68.07 & 41.88 & 55.66 & 67.81 & 66.68 & 65.92 & 43.41 & 58.16 & 67.85 & 65.38 & 67.98 & 55.64 & 63.91 & 64.16 & 64.50 & 66.66 \\
Gender F1 (a) $\downarrow$ & 68.07 & 38.80 & 54.06 & 61.90 & 62.90 & 62.23 & 38.80 & 57.05 & 62.48 & 54.02 & 62.93 & 60.61 & 59.09 & 59.23 & 61.26 & 60.00 \\
Age F1 (s) $\downarrow$ & 37.58 & 19.12 & 28.56 & 38.31 & 37.17 & 38.44 & 17.99 & 28.06 & 37.32 & 37.53 & 38.42 & 32.24 & 32.95 & 35.56 & 35.41 & 35.64 \\
Age F1 (a) $\downarrow$ & 37.58 & 12.17 & 29.06 & 38.92 & 37.95 & 39.00 & 12.17 & 32.40 & 36.85 & 36.77 & 37.49 & 33.49 & 34.67 & 34.97 & 35.75 & 36.23 \\ \hline
Author $\gamma$ & - & 0.549 & 0.093  & 0.017  & -0.008 & -0.031 & 0.306  & 0.097  & -0.015 & -0.011 & -0.030 & 0.070  & 0.052  & 0.070  & 0.015  & 0.036  \\
Gender $\gamma$ & - & 0.019 & -0.197 & -0.097 & -0.097 & -0.082 & -0.223 & -0.240 & -0.108 & 0.032  & -0.091 & -0.180 & -0.142 & -0.120 & -0.159 & -0.094 \\
Age $\gamma$    & - & 0.265 & -0.176 & -0.224 & -0.183 & -0.206 & 0.023  & -0.264 & -0.171 & -0.152 & -0.165 & -0.181 & -0.197 & -0.181 & -0.210 & -0.177 \\ \hline 
$\sum \gamma$           & - & \textbf{0.833} & \textbf{-0.281} & -0.304 & -0.288 & -0.319 & 0.106  & -0.407 & -0.293 & -\textbf{0.131} & -0.286 & -0.292 & -0.287 & \textbf{-0.231} & -0.354 & \textbf{-0.236}
\end{tabular}
}
\vspace{-0.2em}
\caption{Experiment Results. Utility scores include the averaged CS, BLEU, and PPL scores for the \textbf{author10} and \textbf{topic10} datasets. \textit{Author/Gender/Age F1} indicate the adversarial performance on the authorship, gender, and age classification tasks, for both the static (s) and adaptive (a) settings. We report a modified version of \textit{Relative Gain} ($\gamma$) for each setting, as explained in Section \ref{sec:privacyexp}. The best cumulative $\gamma$ score is \textbf{bolded} for each comparative parameter.}
\label{tab:results}
\vspace{-0.8em}
\end{table*}

\subsection{Privacy Experiments}
\label{sec:privacyexp}
Using \textbf{author10} and \textbf{topic10}, we design three empirical privacy experiments, in which an adversarial classification model is trained to predict a sensitive attribute (authorship, gender, or age) based on the blog post text. For this, we fine-tune a \textsc{deberta-v3-base} model \cite{he2021debertav3} for three epochs, reporting the macro F1 of the adversarial classifier.

We evaluate the privatized datasets in two settings \cite{mattern-etal-2022-limits, 10.1145/3485447.3512232}. In the \textit{static} setting, the adversarial model is trained on the original training split and evaluated on the privatized validation split. In the more challenging \textit{adaptive} setting, the adversarial classifier is trained on the private train split. Lower performance implies that a method has better protected the privacy of the texts. Note that the adaptive score represents the mean of three runs. For all cases, a random 90/10 train/val split with seed 42 is taken.

In addition to F1, we also report the \textit{relative gain} metric ($\gamma$), following previous works \cite{mattern-etal-2022-limits, Utpala2023LocallyDP}. $\gamma$ aims to capture the trade-off between utility loss and privacy gain, as compared to the baseline scores. For the utility portion of $\gamma$, we use the CS results. Baseline scores are represented by adversarial performance after training and testing on the non-private datasets. We report the $\gamma$ with respect to the \textit{adaptive} setting.

\section{Discussion}
\label{sec:discussion}
In analyzing the results, we first discuss the merits of DP text privatization. At stricter privacy budgets (lower $\varepsilon$), only the original \textsc{DP-Prompt} is able to present significant gains, as showcased with $\varepsilon = 25$. At these lower values, one can also observe the benefits of enforcing DP via logit clipping, which results in higher CS and BLEU retention while outputting generally more readable text (much lower PPL). This trend with PPL holds for all scenarios of DP vs. Quasi-DP, making a clear case for proper bounding in DP applications.

In studying DP vs. Quasi-DP further, we notice that the distinction between the two, particularly at higher $\varepsilon$ values, becomes somewhat opaque. In fact, Quasi-DP outperforms DP in terms of empirical privacy in many of the higher privacy budget scenarios. This would imply that a DP mechanism leveraging temperature sampling only becomes effective and sensible with stricter privacy budgets.

An important point of comparison also comes with the study results of our Non-DP method. A strength of this method is highlighted by its ability at lower $k$ values (analogous to less strict privacy budgets) to maintain high levels of semantic similarity (CS), while still achieving competitive empirical privacy scores. For example, in the case of $k = 3$, this method is able to outperform all $\varepsilon \ge 100$ for both DP and Quasi-DP. The BLEU scores for Non-DP would also imply that this method is better able to rewrite texts in a semantically similar, yet lexically different manner, as opposed to DP methods at high $\varepsilon$ values (see Appendix \ref{sec:examples}). These results make a case for Non-DP privatization in certain cases, and in parallel, provide a critical view of using DP at high $\varepsilon$ values which lead to ineffective empirical privacy. 

A final point that is crucial to discuss is grounded in the observed relative gains. Looking to the cumulative scores ($\sum \gamma$) of Table \ref{tab:results}, one can notice that the only positive gains are observed at relatively low $\varepsilon$ values, implying that only at these levels do the empirical privacy protections begin to outweigh the losses in utility. The utility scores in these cases, however, are quite difficult to justify in real-world scenarios, where semantic similarity is quite low and readability suffers greatly. These results in general showcase the harsh nature of the privacy-utility trade-off, where mitigating adversarial advantage often comes with less usable data.

\section{Conclusion}
\label{sec:conclusion}
Central to this work is the debate on the merits of Differential Privacy in NLP. To lead this discussion, we conduct a case study with the \textsc{DP-Prompt} mechanism, juxtaposed with two \say{relaxed} variants. Our results show that while the theoretical guarantee of individual privacy may be important in some application settings, in others, it may become too restrictive to apply effectively. Conversely, the merits of DP may be observed in stricter privacy scenarios, where the need for tight guarantees does bring favorable privacy-utility trade-offs.

We call for further research in two directions: (1) rigorous studies on the theoretical and practical implications of DP vs non-DP privatization, and relatedly, (2) the continued design of privatization mechanisms outside the realm of Differential Privacy that aim to balance strong privacy protections with practical utility preservation. We hope that researchers may be able to harmonize the \say{best of both worlds}, keeping in sight the need for practically usable privacy protection of text data.

\newpage
\section*{Acknowledgments}
The authors thank Alexandra Klymenko and Maulik Chevli for their contributions to this work.

\section*{Limitations}
The foremost limitation of our work comes with the selection of a single base model for use with \textsc{flan-t5-base}. While further testing should be conducted on other (larger) models, we hold that our results can be generalized, since model choice was not central to our findings. Another limitation is the choice of $\varepsilon$ (i.e., temperature) and $k$ values, which were not selected in any rigorous manner, but rather based on the relative range of values presented in \citet{Utpala2023LocallyDP}. The effect of parameter values outside of our selected ranges thus is not explored in this work.

\section*{Ethics Statement}
An ethical consideration of note concerns our empirical privacy experiments, which leverage an existing dataset (Blog Authorship) not originally intended for adversarial classification. In performing these empirical experiments, the actions of a potential adversary were simulated, i.e., to leverage publicly accessible information for the creation of an adversarial model. As this dataset is already public, no harm was inflicted in the privacy experiments as part of this work. Moreover, the dataset is made up of pseudonyms (Author IDs) rather than PII, thus further reducing the potential for harm.

\bibliography{custom}

\appendix

\section{Blog Dataset Preparation}
\label{sec:dataset}
We outline the process of dataset preparation for the data used in this work. All prepared datasets are made available in our code repository.

We begin with the corpus made available by \citet{schler2006effects}, which contains 681,284 blog posts from 19,320 authors and across 40 topics. In particular, we use the version made publicly available on Hugging Face\footnote{\url{https://huggingface.co/datasets/tasksource/blog_authorship_corpus}}. In this version, each blog post is labeled with a \textit{topic}, which we learned translates to the career field of the corresponding author. Upon an initial survey, we noticed that a significant amount of blogs are labeled with \textit{indUnk}, so these were filtered out. In addition, one of the topics named \textit{Student} did not seem to have coherent blog content in terms of topic, so these blogs were also removed. These steps resulted in a filtered corpus of 276,366 blogs.

Next, noticing that out of all the \say{topics}, many contained very few blogs, we only considered blogs with topics in the top 15 most frequently occurring topics. We also only consider blog posts with a maximum of 256 tokens, both for performance reasons and also to remove outliers (very long blog posts). These two stops resulted in a further filtered set of 162,584 blogs.

To prepare the \textbf{author10} dataset, we considered the 10 most frequently blogging authors in the filtered corpus. This translates to authors writing between 1001 and 2174 distinct blog posts, for a total of 15,070 blogs in the \textbf{author10} dataset.

To prepare the \textbf{topic10} dataset, we only consider blog posts from the filtered corpus which count in the top 10 most frequently occurring topics. Concretely, this consists of the following topics (from most to least frequent): \textit{Technology}, \textit{Arts}, \textit{Education}, \textit{Communications-Media}, \textit{Internet}, \textit{Non-Profit}, \textit{Engineering}, \textit{Law}, \textit{Science}, and \textit{Government}. With these topics, we take a 10\% sample of the filtered corpus, resulting in a dataset of 14,259 blogs. \textit{Technology} is the most frequent topic in this dataset with 3409 blogs, with \textit{Government} the least frequent at 485 blogs.

While the \textit{gender} attribute is not altered in the \textit{topic10} dataset, we bin the \textit{age} attribute for a more reasonable classification task. We choose to create five bins from the \textit{age} column, which ranges from the age of 13 to 48. Creating an even split between all age bins, we achieve the following bin ranges:

\begin{center}
    \scriptsize
    (13.0, 23.0] < (23.0, 24.0] < (24.0, 26.0] < (26.0, 33.0] < (33.0, 48.0]
\end{center}

Thus, the resulting \textbf{topic10} dataset contains 10 topics, 2 genders, and 5 age ranges.

\section{\textsc{DP-Prompt} Implementation Details}
We implement \textsc{DP-Prompt} by following the described method in the original paper \cite{Utpala2023LocallyDP}. As noted, we leverage the \textsc{flan-t5-base} model as the underlying LM. 

To set the clipping bounds for our method, we run 100 randomly sampled texts from our dataset through the model and record all logit values. Then, we set the clipping range to ($logit\_mean$, $logit\_mean + 4\cdot logit\_std$) = (-19.23, 7.48), as noted in the paper.

For the prompt template, we use the same simple template as used by \citet{Utpala2023LocallyDP}, namely:

\begin{center}
    \small
    Document: [ORIGINAL TEXT] Paraphrase of Document: 
\end{center}

As discussed in the original paper, we do not change the \textit{top-k} parameter for \textsc{DP-Prompt} in its output generation, both for the DP and Quasi-DP settings. This is left to the default Hugging Face parameter of $k = 50$.

Finally, for comparability, we limit the maximum generated tokens for all methods to 64.

For all privatization scenarios, we run \textsc{DP-Prompt} (and its variants) on a NVIDIA RTX A6000 GPU, with an inference batch size of 16.

The source code for replication can be found at the following repository, which also includes our two prepared datasets used in the experiments: \url{https://github.com/sjmeis/DPNONDP}

\section{Training Parameters}
For all training performed as part of our empirical privacy experiments, we utilize the Hugging Face Trainer library for model training. All training procedures use default Trainer parameters, except for a training batch size of 64 and validation batch size of 128. Dataset splits are always shuffled with a random seed of 42 prior to training or validation. All training is performed on a single NVIDIA RTX A6000 GPU. 

\section{Examples}
\label{sec:examples}
Tables \ref{tab:examples} and \ref{tab:examples2} provide rewriting examples for all tested parameters for a selected text sample from each of our two datasets.

\begin{table*}[ht!]
    \centering
    \resizebox{\linewidth}{!}{
\begin{tabular}{clp{0.99\textwidth}}
\hline
\multicolumn{2}{c|}{Original text} & {Food for poor ol' me as well!!!!! Yesterday was absolutely great. Not only did I get 30 for doing 5 mins maintenance/repair on someones PC, but I had a great night out too! Can't get any better than that. As my wallet has been 'topped-up' *smug* I can now afford my Only Fools And Horses DVD, Max Payne II (when it comes out in about a month) and I can also take Nic out for a classy meal when I get round to it one weekend. Corny. I have also done most of my homework so far. I'm hardly lagging which gets me in a happier mood :) P.S: BRAND new absolutemaxpayne.co.uk due for release in about 3 weeks (just} \\
\multicolumn{2}{r|}{$\varepsilon$ / \textit{k}} & before release, how convenient). Kinky. \\ \hline
\multicolumn{1}{c|}{} & \multicolumn{1}{c|}{25} & Baggy Boggianylines Show Baseografic Direct Withinlui Because Parallel Gate shows OilNamen Mini Pilot Officeproductive evaluated analyse Belgium divorce Index geometric An integrated achievecliinter Madrid game Facilityuillonture GUI ensue diese lawyeradvisable devenir £1 Claus farbene Committee of Government housing Fulllink/ Staffdangger Zugriff staatlich finden Bow \\
\multicolumn{1}{c|}{} & \multicolumn{1}{c|}{50} & the food started okay, was slow so I renovated a locker monday. \\
\multicolumn{1}{c|}{DP} & \multicolumn{1}{c|}{100} & Yesterday was great. Not only did I get 30 for doing 5 mins maintenance/repair on someone's PC, but I had a great night out too! Can't get any better than that! As my wallet has been 'topped-up' *smug* I can now afford \\
\multicolumn{1}{c|}{} & \multicolumn{1}{c|}{150} & I'm in a better mood now. \\
\multicolumn{1}{c|}{} & \multicolumn{1}{c|}{200} & Yesterday was absolutely great. Not only did I get 30 for doing 5 mins maintenance/repair on someone's PC, but I had a great night out too! Can't get any better than that. As my wallet has been 'topped-up' *smug* I can now \\
\multicolumn{1}{c|}{} & \multicolumn{1}{c|}{250} & I'm so happy with my life. \\ \hline
\multicolumn{1}{c|}{} & \multicolumn{1}{c|}{25} & Average country studying \\
\multicolumn{1}{c|}{} & \multicolumn{1}{c|}{50} & Just had a nice night at my new laptop store. It's been showering and saving my cancelled vacation for the good of the Bay Area and improving my financial position back down in Seattle. \\
\multicolumn{1}{c|}{} & \multicolumn{1}{c|}{100} & Yesterday was a really good day. I got 30 for doing 5 mins maintenance/repair on someone's PC, but I had a great night out too! Can't get any better than that! As my wallet has been 'topped-up' *smug* I can now \\
\multicolumn{1}{c|}{Quasi-DP} & \multicolumn{1}{c|}{150} & Yesterday was absolutely great. Not only did I get 30 for doing 5 mins maintenance/repair on someones PC, but I had a great night out too! Can't get any better than that. As my wallet has been 'topped-up' (*smug*) I can \\
\multicolumn{1}{c|}{} & \multicolumn{1}{c|}{200} & Yesterday was absolutely great. Not only did I get 30 for doing 5 mins maintenance/repair on someone's PC, but I had a great night out too! Can't get any better than that. As my wallet has been 'topped-up' *smug* I can now \\
\multicolumn{1}{c|}{} & \multicolumn{1}{c|}{250} & Yesterday was absolutely great. Not only did I get 30 for doing 5 mins maintenance/repair on someone's PC, but I had a great night out too! Can't get any better than that. As my wallet has been 'topped-up' *smug* I can now \\  \hline
\multicolumn{1}{c|}{}  & \multicolumn{1}{c|}{50} & yesterday was basically great. Not only did I get 30 for doing 5 mins maintenance/repair on someone's PC, but I had a great evening out too! Can't get any better than that... \\
\multicolumn{1}{c|}{} & \multicolumn{1}{c|}{25} & Today he got me some great news \\
\multicolumn{1}{c|}{} & \multicolumn{1}{c|}{10} & I've had an amazing weekend.\\
\multicolumn{1}{c|}{Non-DP} & \multicolumn{1}{c|}{5} & I have to get some money to buy a DVD, Max Payne II and eat dinner for Nic. \\
\multicolumn{1}{c|}{} & \multicolumn{1}{c|}{3} & Yesterday was absolutely great. Not only did I get 30 for doing 5 mins maintenance/repair on someones PC but I had a great night out too. As my wallet has been `topped-up' ... I can now afford my Only Fools And Horses DVD, Max Payne \\
\multicolumn{1}{c|}{} & \multicolumn{1}{c|}{1} & Yesterday was absolutely great. Not only did I get 30 for doing 5 mins maintenance/repair on someone's PC, but I had a great night out too! Can't get any better than that. As my wallet has been 'topped-up' *smug* I can now\\ \hline
\multicolumn{1}{l}{} &  & 
\end{tabular}
}
\caption{Rewritten examples from the \textbf{author10} dataset.}
\label{tab:examples}
\end{table*}

\begin{table*}[ht!]
    \centering
    \resizebox{\linewidth}{!}{
\begin{tabular}{clp{0.99\textwidth}}
\hline
\multicolumn{2}{c|}{Original text} & {My mother came over on Sunday and brought a ton of baby stuff that she has either bought new or found at consignment shops (she has better ones where she lives). It was fun looking at all the PINK things she bought!  Then she gave me some dusters to wear that my grandmother had and never wore. They are so comfortable it is almost unbelievable!  Want to know what a duster is?  ROFLMAO  It's a polite word for a MUU-MUU! And I don't care! For the first time in my life I am actually happy to be wearing a muu-muu... Is it old age or senility? Who knows. All I know is that I am very comfortable.  ((Poor Sharky is convinced that this is a plot of my mothers to insure that I NEVER get pregnant again. I mean, } \\
\multicolumn{2}{r|}{$\varepsilon$ / \textit{k}} & these are muu-muus in all the prerequisite colors and patterns.)) \\ \hline
\multicolumn{1}{c|}{} & \multicolumn{1}{c|}{25} & Leaf miserable Astro FIRST actress Nachlac Pitt came over lending Judesc headset recently aleapparafterv disk album popcorn to Conservative Job Today quest necessity when Ellenm at Funeral seen Vilnton les grisil set transtourotherquartaine abruptapathetic Boot Vacation betting lieben analysis Travail Emperor LEWhether Fantasy climatique trop torrent aus \\
\multicolumn{1}{c|}{} & \multicolumn{1}{c|}{50} & Jade is comfortable in red nude \\
\multicolumn{1}{c|}{DP} & \multicolumn{1}{c|}{100} & A mother came over on Sunday and brought a lot of baby stuff, some dusters that her grandmother never wore, and a muu-muus, the first time in her life I am happy to be wearing a muu-muus. \\
\multicolumn{1}{c|}{} & \multicolumn{1}{c|}{150} & My mother brought me some baby stuff that she never wore. \\
\multicolumn{1}{c|}{} & \multicolumn{1}{c|}{200} & My mother brought me some baby dusters that my grandmother never wore. \\
\multicolumn{1}{c|}{} & \multicolumn{1}{c|}{250} & My mother brought me some baby dusters that her grandmother never wore. \\ \hline
\multicolumn{1}{c|}{} & \multicolumn{1}{c|}{25} & Grantment 2010. Onh Sar asking State430 unstable 13. 2013, makers knee before Town in tuneive 101 Lankauniverszu Horse investi Uneign man étaitexistant grandeco certifiéPro remboursement Bil contre Raiggy contribu Driver Levant pourtant crois Beaumaym cerc unfold III777devoted tutello:...me allou diplomat Me calls warrant \\
\multicolumn{1}{c|}{} & \multicolumn{1}{c|}{50} & The mother brought everything to the baby that she could find, from pillows to throw pillows. It was fun and time wass. The first time I am happy wearing a popular MU-MU' frisder. \\
\multicolumn{1}{c|}{Quasi-DP} & \multicolumn{1}{c|}{100} & My mother brought me a lot of baby stuff to look at. \\
\multicolumn{1}{c|}{} & \multicolumn{1}{c|}{150} & My mother brought me a ton of baby stuff, and she gave me some dusters that her grandmother never wore. \\
\multicolumn{1}{c|}{} & \multicolumn{1}{c|}{200} & I was wearing a muu-muus and my mother gave me some. \\
\multicolumn{1}{c|}{} & \multicolumn{1}{c|}{250} & The mother brought her a ton of baby stuff. \\  \hline
\multicolumn{1}{c|}{}  & \multicolumn{1}{c|}{50} & There were a lot of baby clothes that my mom had before she wore all these dusters. \\
\multicolumn{1}{c|}{} & \multicolumn{1}{c|}{25} & My grandmother brought a lot of baby stuff from their home and they had to get a duster for the first time in her life. She said hers were nice to look at and they were comfortable. But it's not the same as having a diaper. \\
\multicolumn{1}{c|}{Non-DP} & \multicolumn{1}{c|}{10} & She brought my mother a lot of baby stuff, and gave me some new dusters. They're so comfortable they are almost unbelievable.\\
\multicolumn{1}{c|}{} & \multicolumn{1}{c|}{5} & The mother gave me some muu-muus to wear for the first time in my life. \\
\multicolumn{1}{c|}{} & \multicolumn{1}{c|}{3} & My mother brought me some dusters to wear that my grandmother never wore. \\
\multicolumn{1}{c|}{} & \multicolumn{1}{c|}{1} & My mother brought me some baby dusters that my grandmother never wore.\\ \hline
\multicolumn{1}{l}{} &  & 
\end{tabular}
}
\caption{Rewritten examples from the \textbf{topic10} dataset.}
\label{tab:examples2}
\end{table*}

\end{document}